\documentclass{article}

\usepackage{PRIMEarxiv}

\usepackage[utf8]{inputenc} 
\usepackage[T1]{fontenc}    
\usepackage{hyperref}       
\usepackage{url}            
\usepackage{booktabs}       
\usepackage{amsfonts}       
\usepackage{nicefrac}       
\usepackage{microtype}      
\usepackage{lipsum}
\usepackage{fancyhdr}       
\usepackage{graphicx}       
\graphicspath{{media/}}     
\usepackage{makecell}
\usepackage{colortbl}
\usepackage{booktabs}
\usepackage{multirow}
\usepackage{hhline}
\usepackage{amssymb}
\usepackage{pifont}
\usepackage{verbatim}
\newcommand{\cmark}{\ding{51}}%
\newcommand{\xmark}{\ding{56}}%
\definecolor{Gray}{gray}{0.8}
\definecolor{LightGray}{gray}{0.95}
\definecolor{ForestGreen}{rgb}{0.13, 0.55, 0.13}
\definecolor{White}{rgb}{1.0, 1.0, 1.0}
\usepackage[bottom]{footmisc}

\title{Joint Embedding of 2D and 3D Networks\\for Medical Image Anomaly Detection
}

\author{
  Inha Kang \\
  KAIST \\
  \texttt{rkswlsj13@kaist.ac.kr} \\
   \And
  Jinah Park\thanks{corresponding author} \\
  KAIST \\
  \texttt{jinahpark@kaist.ac.kr} \\
}

\usepackage{pdfpages}
\begin{document}
\maketitle

\begin{abstract}
Obtaining ground truth data in medical imaging has difficulties due to the fact that it requires a lot of annotating time from the experts in the field. Also, when trained with supervised learning, it detects only the cases included in the labels. In real practice, we want to also open to other possibilities than the named cases while examining the medical images. As a solution, the need for anomaly detection that can detect and localize abnormalities by learning the normal characteristics using only normal images is emerging. With medical image data, we can design either 2D or 3D networks of self-supervised learning for ‘Anomaly Detection’ task. Although 3D networks, which learns 3D structures of the human body, show good performance in 3D medical image anomaly detection, they cannot be stacked in deeper layers due to memory problems. While 2D networks have advantage in feature detection, they lack 3D context information. In this paper, we develop a method for combining the strength of the 3D network and the strength of the 2D network through joint embedding. We also propose the pretask of self-supervised learning to make it possible for the networks to learn efficiently. Through the experiments, we show that the proposed method achieves better performance in both classification and segmentation tasks compared to the SoTA method.
\end{abstract}


\section{Introduction}

\begin{figure}[!t]
    \vspace{-1em}
    \centerline{\includegraphics[width=15cm]{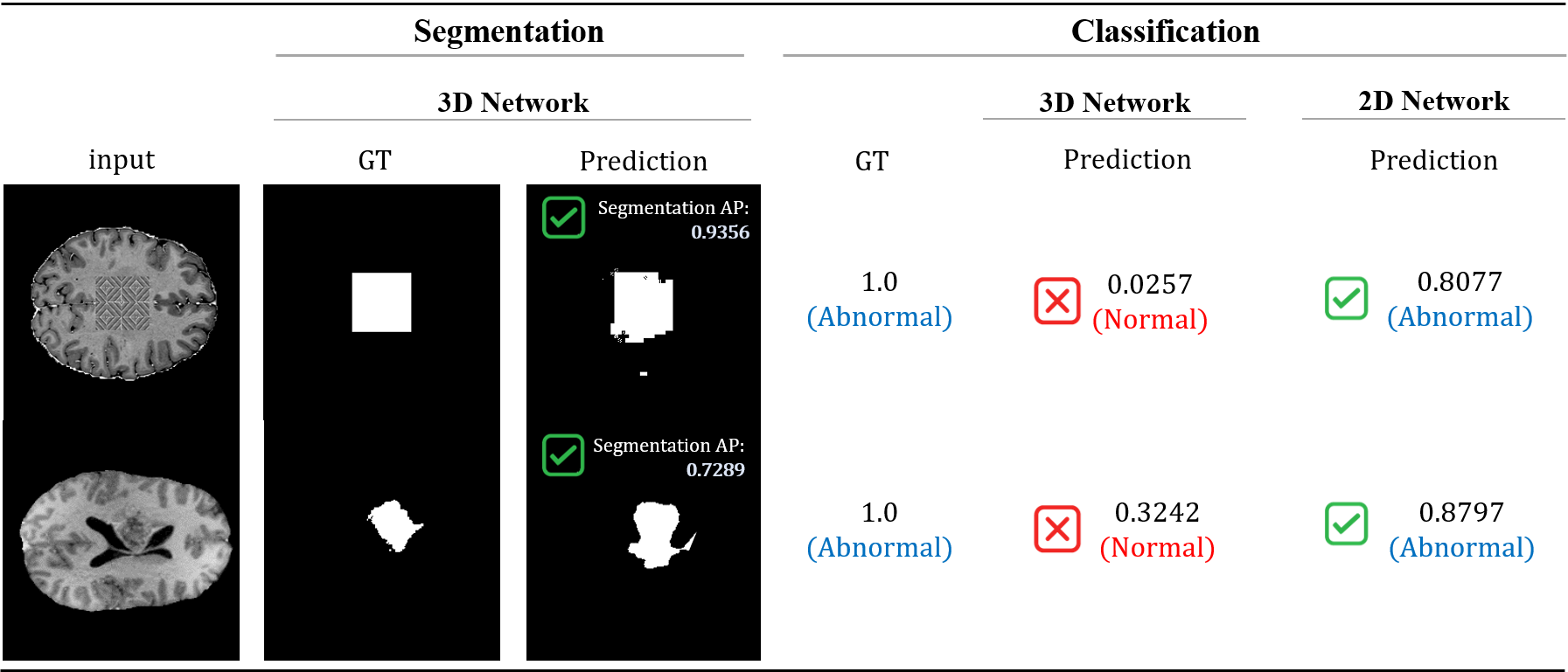}}
    \caption[Results of the 3D SoTA Network and the 2D ResNet-50 Network.]{\textbf{Results of the 3D SoTA Network and the 2D ResNet-50 Network. }It shows the average precision results of each network. The 3D U-Net based model shows good segmentation results, but not classification results. However, the ResNet-50 model classify accurately for the same image.
    } \label{intro}
\end{figure}
\begin{figure}[!t]
    \centerline{\includegraphics[width=16.5cm]{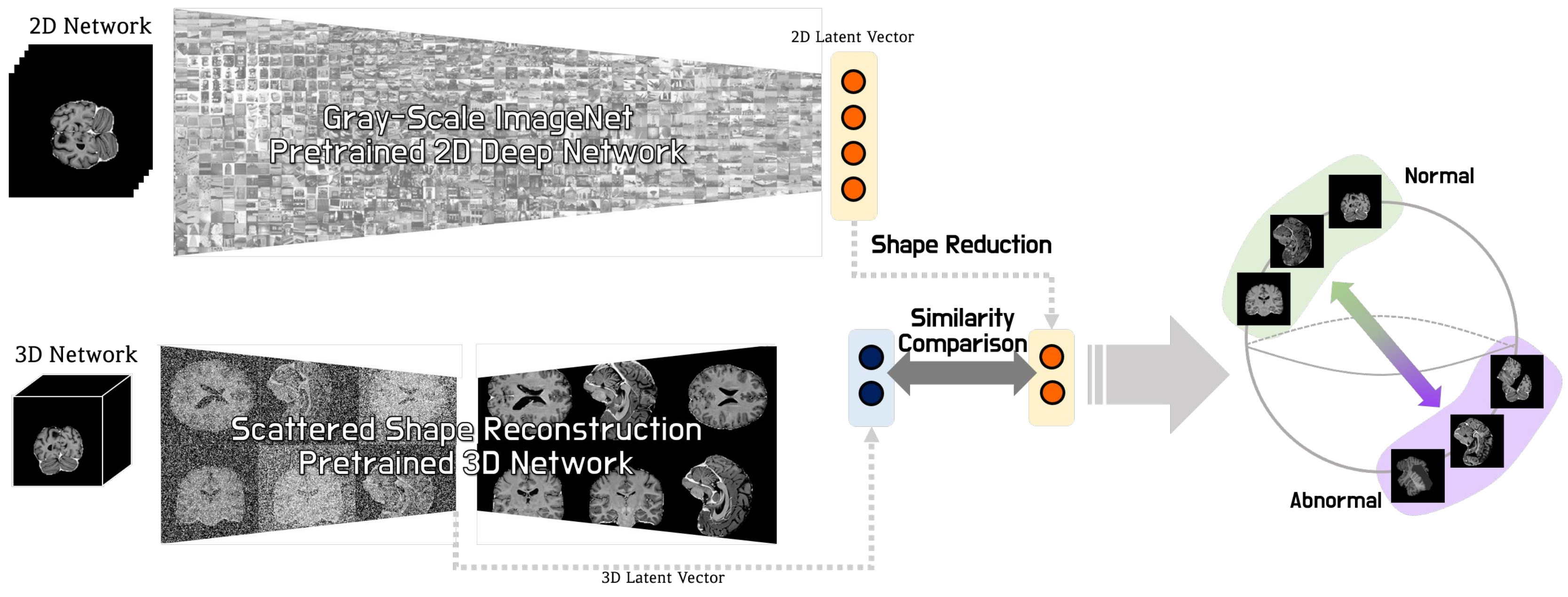}}
    \caption[Overall Scheme of the Proposed Methods.]{\textbf{Overall Scheme of the Proposed Methods. }2D deep networks are pretrained with grayscale large datasets, and 3D volumetric networks are pretrained with scattered shape reconstructions. The latent vectors of each network are compared for their similarity.
    } \label{abstract}
\end{figure}

\noindent
As the performance of deep learning has greatly improved, studies using deep learning as an auxiliary diagnosis are being widely used in the medical imaging field. The need for a secondary diagnostic tool is getting more attention following up the fact that even experts in the field can commit some mistakes or miss important points present in the image data. When the radiologists focused on the lung nodule for the image synthesized from the gorilla image into the CT scan, 83\% of the radiologists did not detect the gorilla~\cite{gorilla}. It shows that even well-trained radiologists often fail to detect abnormalities in unexpected situations. Such mistakes may be fatal since medical image analysis is directly related to human life. On the other hand, generating labels for medical image data with various types of lesions or tumors is non-trivial and expensive task since it requires a considerable amount of time by the experts in the field. Moreover, even if supervised learning is enabled with labeled images, it is hard to find a type of lesion or tumor that is not included in the training dataset. Therefore, it is necessary to devise an unsupervised learning-based network that learns the distribution of normal data so that classify Out-Of-Distribution (OOD) data. \\
In the field of medical imaging, there are many 3D modalities such as CT or MRI. For 3D images, many studies have demonstrated that 3D networks that can learn 3-dimensional context outperform 2D networks ~\cite{brats, moodchallenge}. 3D networks, however, have limitations on depth of network and lack of datasets. In various encoder-decoder structures based on 3D U-Net, it shows good performance in segmentation task, but relatively low performance in classification tasks as shown in Figure ~\ref{intro}. It reflects shortages of a 3D network that has relatively shallow classification modules due to its memory limitation. 2D networks, on the other hand, can use much deeper networks with less memory than 3D networks and do not significantly boost memory usage even for high-resolution inputs. In addition, unlike the medical imaging field, where datasets are scarce, there are many natural image datasets that can be used as pretraining weights in 2D network training. We attempt online joint embedding of 2D and 3D networks as in Figure \ref{abstract}. Latent vectors of 2D networks are pretrained on large grayscale natural images, which helps to extract meaningful features~\cite{pretrain1, pretrain2, pretrain3}. Similarity loss is used for joint embedding between 2D network and 3D network. It is updated by an online learning method to align representations in embedding space.  \\

Additionally, we use the pretrained 2D model with grayscale ImageNet-1k~\cite{imagenet} and the scatter-shaped reconstruction pretask in the 3D model. Pretraining and pretask help fast training by solving the problem of joint embedding that it takes too long to converge. Therefore, fast embedding is induced by applying self-supervised pretasks rather than training 3D network from scratch. According to experiments on various self-supervised pretasks, scattered reconstruction shows the highest performance. It verifies that convergence and performance are improved when fine-tuning the result of pretask to the 3D model. \\
\indent We propose joint embedding of two networks that have different dimensions. Before the anomaly detection training phase, we apply pretraining to the 2D network which utilizes large amount of natural image dataset and pretask to the 3D network for fast convergence. As a result, our method  detects small brain tumor or lesions that are difficult to distinguish due to similar intensity values without long training time, unlike 3D U-Net based models that cannot detect them. \\
\vspace{-1.5em}
\paragraph{Contributions of the thesis:}  
\begin{itemize}
\vspace{-.5em}
    \item We propose a new idea on joint embedding of 2D and 3D networks through similarity loss, allowing us to simultaneously take advantages of 2D and 3D networks in the medical imaging field.
    
    \item We apply the large-scale natural image pretraining in the 2D network and the scattered-shaped reconstruction pretask in the 3D network to stimulate fast joint learning.
    
    \item We show SoTA results on the medical image anomaly detection dataset, MOOD 2021.
\end{itemize}

\section{Related Works}

\subsection{Self-Supervised Learning}
\textbf{Pretask. }Self-Supervised Learning aims to learn the inherent features of data without manual annotation. The network is trained to perform a specific task that does not needs any human supervision. Through a pretask step, the network learns the overall features of the dataset although its task is not the same as the original task. Various types of pretasks are being used for self-supervised learning, such as colorization, solving jigsaw puzzles and context prediction. Recently, a lot of transformer models \cite{transformer} based on natural language processing are being studied, so some papers \cite{beit, masked} show remarkable results by inpainting as a pretask with tokenized images. With these novel approaches, nowadays other researches related on reconstructing tokenized images are actively conducted. Recently, ConvMAE~\cite{convMAE} reveals that the performance difference between various hybrid convolution-transformer backbones and the simple model MAE backbone is not significant. Through this, research is conducted in the direction of optimizing MAE with a new masking strategy and hybrid backbone are introduced.

\noindent \textbf{Contrastive Learning. }Instance discrimination tasks based on contrastive learning are showing state-of-the-art performance in many areas \cite{simclr, simclr2, moco, csi}. SimCLR \cite{simclr} learns to classify different samples by using contrastive loss. It uses multiple data augmentations to generate positive pairs, such as random cropping, color distortions and Gaussian blur. The network aims to maximize agreement between positive pairs in latent space. In the medical image field, there are cases of using contrastive loss by taking advantage of the fact that there are distinguishable features according to the position of the image ~\cite{FCL}. In this paper, after dividing a 3D volume into several sections, slices in the same section are defined as positive pairs, and slices in other sections are defined as negative pairs. CSI \cite{csi} is another self-supervised learning method designed for anomaly detection. In SimCLR and CSI, it is reported that some hard augmentations, which may spoil important features of given image, can degrade the model's discriminate performance. In our previous works, it motivated us to use those hard augmentations, \textit{copy-paste} for pseudoanomaly generation \cite{moodchallenge}.


\subsection{Representation Learning}
\textbf{Knowledge Distillation. }Knowledge Distillation aims to transfer knowledge of pretrained huge networks to relatively light networks. It is firstly devised by Hinton et al~\cite{KD_1}, with the idea that wide and deep networks can extract the key point features well. Therefore, if the huge network, coined as teacher network, can distillate their feature extract performances to small networks, coined as student networks, it can be said as memory-efficient networks. Initially, knowledge distillation models use two loss function, student loss and distillation loss. Student loss refers that general cross entropy loss between student network's prediction results and labels. Distillation loss refers that comparing the results between teacher networks and student networks to distillate the teacher's knowledge to student \cite{KD_2}. Some paper use the knowledge distillation for model compression \cite{KD_3}. Knowledge distillation is viewed as function matching, and when knowledge distillation is applied with the function matching method proposed in this paper, the student reaches the teacher's accuracy. The accuracy of the teacher and the student is the same, but the model size of the student is small, so it can be thought of as a model compression method. By applying knowledge distillation to a model that people lightly use this latest model, we get a generalized model. Distillation for model compression highlights two methods. First, augmentation with the same input image of teacher and student should be applied. The second emphasizes the incredibly long learning times. According to the results of this paper \cite{KD_3}, it can be confirmed that the performance of knowledge distillation takes a extremely long time to train.

\noindent \textbf{Joint Embedding. }Multi-modal learning is a method of using data types with different characteristics rather than one type of data. The difficulty of learning multi-modal data is that they have different representations according to their modalities. For example, audio is a 1D signal wave form, an image can be represented as a 2D or 3D array and the text has embedding data corresponding to the word~\cite{JE1}. The method of learning embedding vectors to match different representations is called joint embedding. The most important aspect of cross-modality research is learning a common subspace in which different types of modalities can be directly compared. At this time, after combining the two pretrained uni-modal models, the mapping is conducted in the same embedding space. For this, the metric training step of learning whether to push or pull the distance between the two distances is performed~\cite{JE2}. Pulling and pushing between representations are often calculated by similarity-based function. Unsupervised learning also can be achieved through the boundary obtained by similarity comparison~\cite{JE3,JE4}. Many studies are interested in which different types of representations generate similar representations in a common subspace and learn in adversarial way. By applying adversarial learning to joint embedding, encoders of different modalities are able to learn competitively to improve accuracy~\cite{JE5}. Unlike knowledge distillation, which trains the student model to follow the performance of the teacher model, joint embedding is learned complementarily by fitting different representations in a common subspace. We think that these points of joint embedding can utilize the advantages of 2D network and 3D network at the same time. Therefore, it is newly proposed to jointly embedding the representation of a 2D network and a representation of a 3D network in one subspace, not in order to solve the problem caused by different modalities, but to take advantage of the advantages of a 2D network and a 3D network.

\begin{figure}[!tb]
    \centerline{\includegraphics[width=16cm]{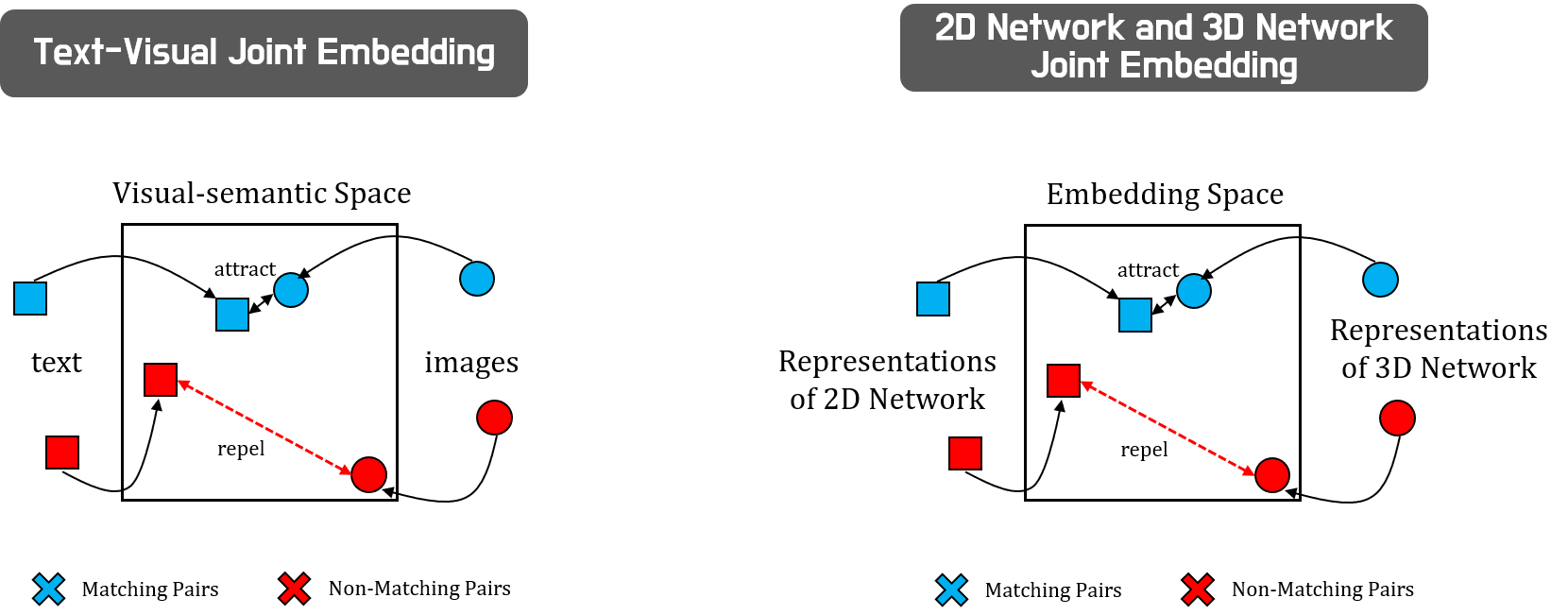}}
    \caption[Examples of Joint Embedding.]{\textbf{Examples of Joint Embedding. }(Left) Joint embedding among different modality, text and images. (Right) Joint embedding between representations of 2D networks and 3D networks, which is newly suggested in this paper.
    } \label{RW_JE}
\end{figure}

\subsection{Anomaly Detection}
\textbf{2D Anomaly Detection. }Anomaly detection, or Out-of-Distribution (OOD) detection can be viewed as classification and localization problems of detecting a sample that comes far from the training data distribution (in-distribution) when only of normal images is provided. The type for anomaly depends on the characteristics of the given data and task. For example, MVTec dataset \cite{MVTec}, which is one of the most frequently used dataset for 2D images anomaly detection, defines anomaly as a variety of defects on the object. It could be scratches, dents, or the absence of certain object parts. Since it is hard to detect anomalies without any normal dataset, many recent studies are focusing on self-supervised learning methods. Patch SVDD \cite{patchsvdd} used patch-based method to project input image into a low-dimensional feature space, in order to emphasize the difference between normal and abnormal samples. CutPaste \cite{scar} is another self-supervised approach that proposed a novel pretext task, which cuts an image patch and pastes at a random location of an image.

\noindent \textbf{3D Anomaly Detection. }In medical imaging field, 3D volumetric datasets are commonly acquired from CT and MRI. However, most of the existing methods to train volume data is trained with 2D slices ~\cite{ssl_2d_med1,ssl_2d_med2}, without taking in much of the 3D context information. Autoencoder-based networks~\cite{autoencoder1, autoencoder2, autoencoder3} were developed, which have the fatal limitation that they cannot be reconstructed to be exactly the same as the original volume. In support of these, many researches say that 3D based networks show better results than 2D networks or autoencoder-based networks \cite{brats, moodchallenge}. In the case of medical image anomaly detection, there are Medical Out-Of-Distribution(MOOD) 2021 dataset~\cite{mood_2021}, which include only normal CT scans of the brain and abdominal. According to the our previous works~\cite{moodchallenge}, which is SoTA of this dataset, utilizes self-supervised methods and 3D U-Net based network with classification module. 3D U-Net \cite{unet} is convolutional neural network architecture for fast and precise segmentation of images, composed of a contracting path to capture context and a symmetric expanding path that enables precise localization. The skip connections that combines the high resolution features from the contracting path to the expanding path help the network to learn a more precise output. It was originally developed for biomedical image segmentation, but nowadays it is showing great performance in a wide range of areas and broadly used in backbone models for various tasks. 

\section{METHOD} \label{sec:sections}

\subsection{Joint Embedding}

\begin{figure}[!h]
    \centerline{\includegraphics[width=17cm]{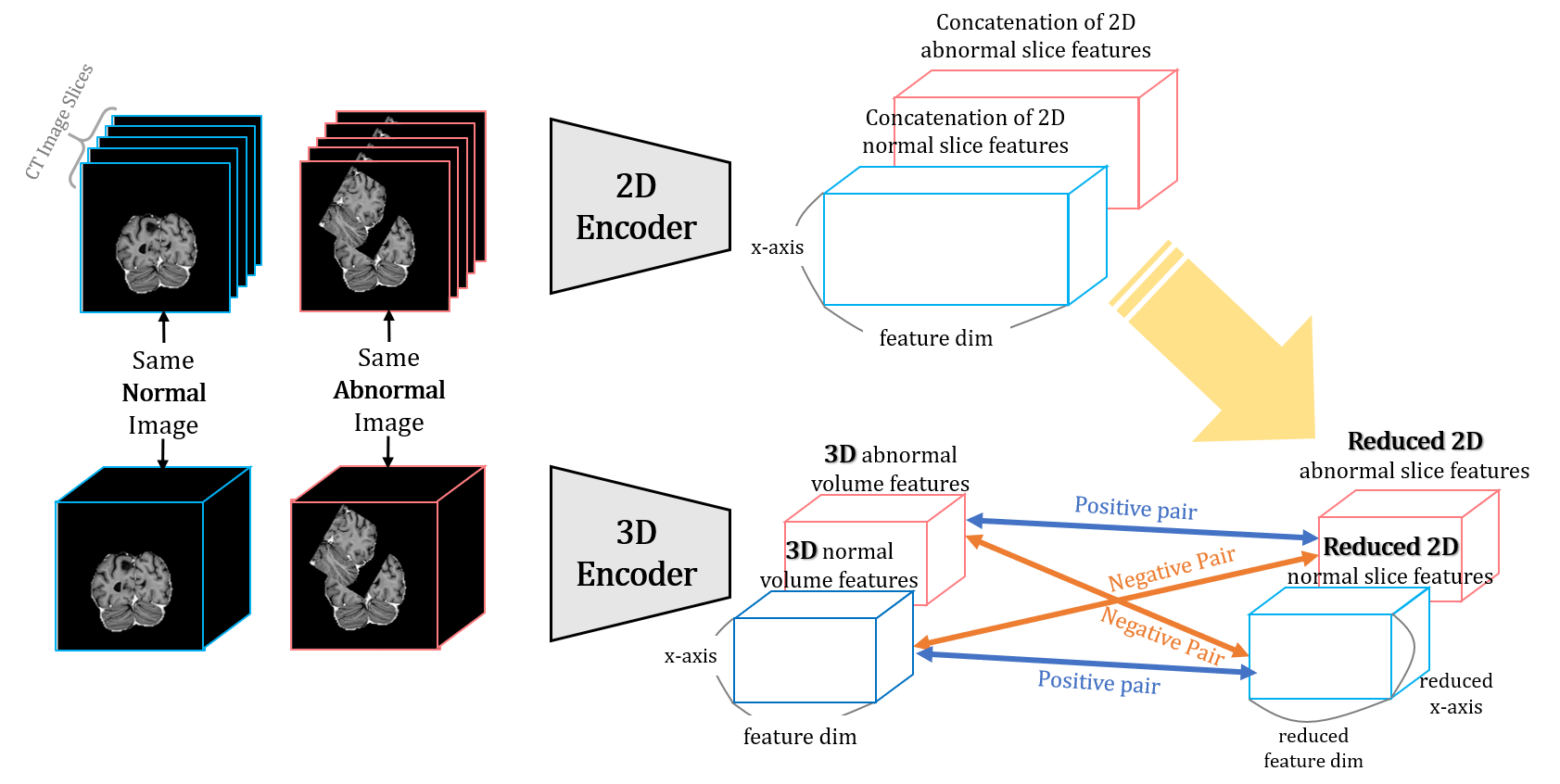}}
    \caption[Overview of the Joint Embedding between the 2D Network and the 3D Network Through Similarity Loss.]{\textbf{Overview of the Joint Embedding between the 2D Network and the 3D Network Through Similarity Loss.} The same color of the latent vectors refers to positive pairs and the different colors of the latent vectors refer to negative pairs. Therefore, there are four cases vector pairs are exist, two positive pairs and two of negative pairs.
    } \label{JE_overview}
\end{figure}
\begin{figure}[!t]
    \centerline{\includegraphics[width=15cm]{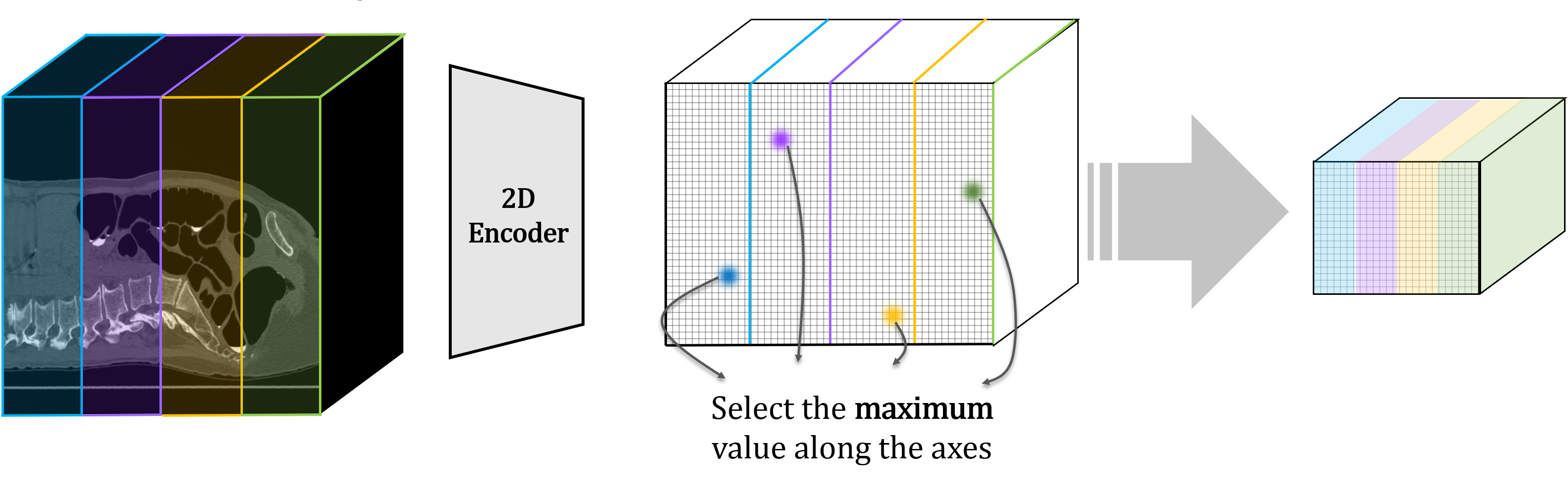}}
    \caption[Method for Reduce the Shape of the 2D Latent Vectors]{\textbf{Method for Reduce the Shape of the 2D Latent Vectors. }2D latent vectors should be reduced along concatenated direction axis and feature dimension axis. In this process, the maximum value of each section becomes representative values.
    } \label{reduction}
\end{figure}

\noindent We present a method that complements the advantages of both 2D networks and 3D networks. Basically, a huge dataset called ImageNet-1k converted to grayscale is trained with a 2D network so that meaningful features can be extracted. In order to utilize the feature extractor which is trained by large-scale dataset with the deep model for the 3D model, a loss that compares the latent vector for the same image is added. The 2D network receives a 2D image slice from the CT image volume as an input, and both models receive a pair of normal image and an anomaly image at the same time so that all features of each can be learned. In this case, there are two positive pairs when the same images are received in each network and two negative pairs are received. In the case of a positive pair, the similarity between the latent vector of the 2D network and the latent vector of the 3D network increases, and in the case of a negative pair, the similarity of the each network can be decreased despite the different network structures by similarity loss. Similarity loss is calculated with cosine similarity.\\

\noindent \textbf{Axis Free Slicing. }When slicing a 3D volume into a 2D image, the 3D volume is rotated and then sliced at a certain axis to prevent learning focused on one axis. By rotating the 3D volume in a random direction and slicing it while training the same normal and abnormal images with the same augmentation to each the 2D network and the 3D network. In this way, the 2D network is trained without focusing on any axis. At this time, the probability of rotation in each direction is set to be the same.

\noindent \textbf{2D Latent Vector Shape Reduction. }In a 2D network, multiple slice images from a 3D volume used as inputs, so in order to match the shape with the 3D latent vector, the latent vectors of the 2D images need to be concatenated. Then, the size of the latent vector in the connected direction increases as much as the number of slices. For this reason, it is necessary to reduce the 2D latent vector. Therefore, we utilize the characteristics of the medical image. Most medical images have unique characteristics according to their position, as can be seen in the Figure~\ref{reduction}. Therefore, after dividing the section by the ratio to be reduced, only the value of the latent vector having the maximum value in each section is left. This is because the result of the bottom latent vector of the 3D network will also retain the most representative features. In this way, only the important information will be updated when compared with the maximum of the 2D latent value by similarity loss. In the same way, for the feature dimension, only the vector with the biggest value is left along the axis of the feature dimension. Through this, the 2D latent vector can be adjusted to the same shape as the 3D latent vector.

\subsection{Pretask}
\begin{figure}[!b]
    \centerline{\includegraphics[width=16cm]{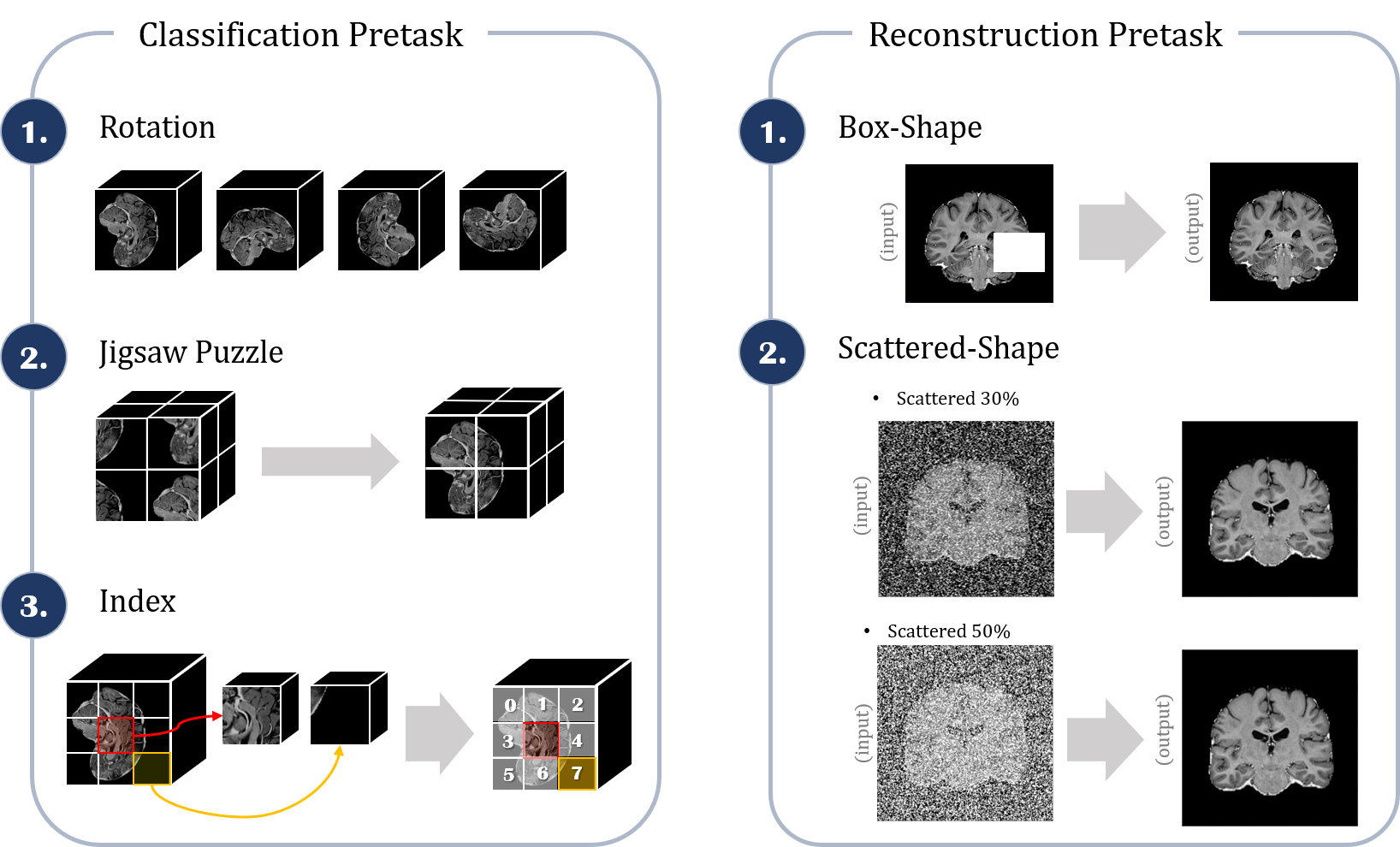}}
    \caption[Examples of Pretasks.]{\textbf{Examples of Pretasks. }Pretasks are divided into two groups, classification tasks and reconstruction tasks. The rotation has 24 cases, the jigsaw puzzle has $8!$ cases and the index has 26 cases. Reconstruction tasks aim to recover missing pixels.
    } \label{pretask}
\end{figure}

\noindent In the case of 2D networks, there are pretrained weights with various large datasets, so training is often performed using them. However, in the case of 3D, since there is no large data set that can be pretrained, most of the training starts from scratch. In this case, not only the training time takes a long time, but the performance is also inferior to the case of using pretraining. Therefore, in order to be able to learn using pretraining weights in 3D networks, we conducted pretask learning before training. By learning in advance through anomaly detection, anomaly segmentation, and other tasks, which were originally intended, we are trying to strengthen the feature extraction ability of 3D networks.\\
\indent It can be divided into a classification-based pretask, which is often used in self-supervised learning, and a newly proposed reconstruction-based pretask. In the case of rotation, it is a task to determine which direction the 3D volume is rotated in. Since it is 3D, it can be rotated in a total of 24 cases. In the case of the jigsaw puzzle, when the 8 pieces obtained after dividing the 3D CT image in half along each axis are mixed, it means to fit the original shape again. It is a task that has as many cases as $8!$. The index task is a classification task that matches the location of a patch when a reference patch is given and a random patch is given among 26 patches around it. Cross entropy loss is used when training the classification pretasks. \\
\indent In general self-supervised learning, most of the pretasks are based on classification. However, since we use the encoder-decoder model, reconstruction-based pretask would be helpful in learning the overall structure of medical images. Box shaped reconstruction refers to the box shaped mask is applied to a certain area of the input image, and it aims to reconstruct same as the original image. Scattered shape refers to the random pixel of input images' within a certain portion that is removed. As in the example in Figure \ref{pretask}, the degree of pixels to be erased is divided into 30\% and 50\% of the total pixels. Only the portion of the pixel being removed is different, and we train the model using $L_{2}$ loss to create the same image as the original image. \\


\subsection{Network Architecture}

\subsubsection{2D Network}
\begin{figure}[!ht]
    \centerline{\includegraphics[width=\columnwidth]{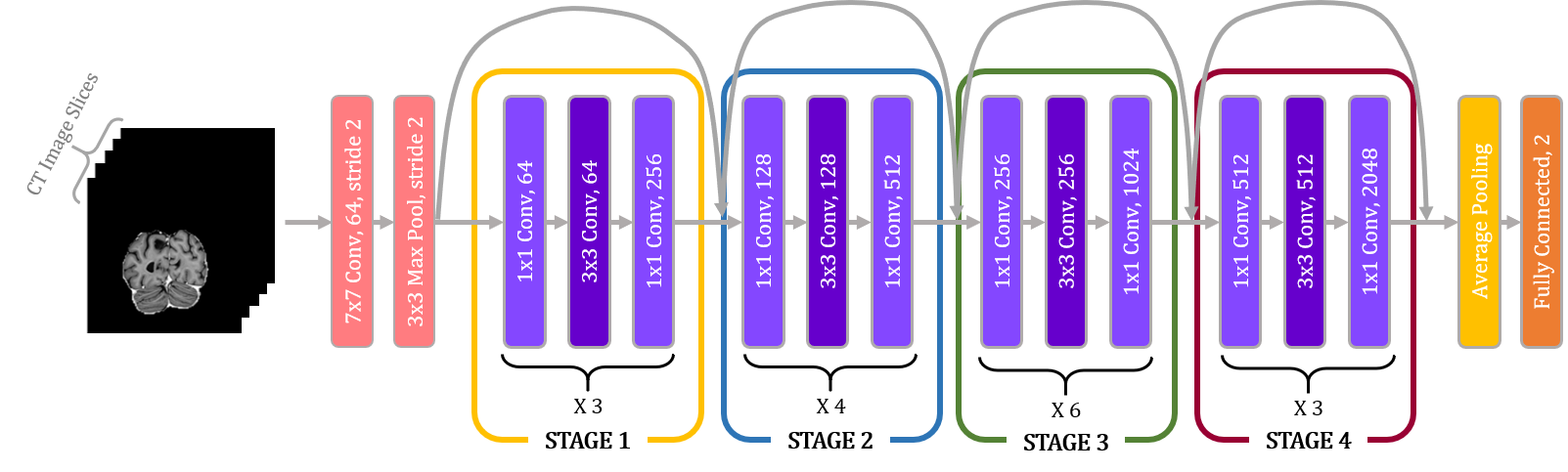}}
    \caption[Network Architecture of the ResNet-50 Model.]{\textbf{Network Architecture of the ResNet-50 Model.} The 2D network is pretrained with gray-scale ImageNet-1k dataset and then takes 2D slices of a 3D volume image.
    } \label{resnet50}
\end{figure}

\noindent \textbf{Pretraining with Gray-scale ImageNet Dataset. }The process of training a deep 2D network such as Figure \ref{resnet50} through ImageNet-1k is conducted first. Medical images are often in gray-scale, so we change the natural images into gray-scale accordingly and then normalized them from 0 to 1. In the case of augmentation, the most basic crop, flip, and rotation are used, and training is proceeded through the Adam optimizer. After training the model for 80 epochs with this large dataset, it is used as a pretrianing weight. In training for anomaly detection, one 3D medical volume is cut into several slices are used as an input image based on the pretrained weight. At this time, since the increase in memory usage according to the image size is not greater than that of a 3D image, a resolution twice as high for the same 3D image is used as the input image. Through this, the advantage of using a higher resolution input image will also contribute to the improvement of accuracy.

\begin{figure}[!h]
    \centerline{\includegraphics[width=6.5cm]{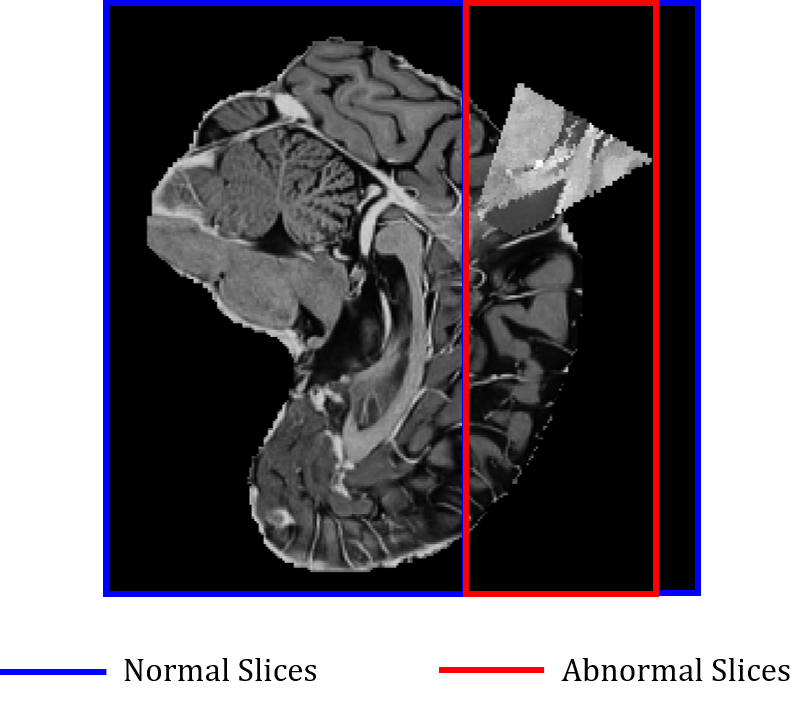}}
    \caption[Class Imbalance Problems of the 2D Network.]{\textbf{Class Imbalance Problems of the 2D Network.} Even if a 3D volume is an anomaly image, all slices are not abnormal. It shows the class imbalance problems of normal slices and abnormal slices.
    } \label{imbalance}
\end{figure}

\noindent \textbf{Weighted Binary Cross Entropy Loss. }When calculating the similarity loss, it is necessary to consider both the normal image and the abnormal image for a positive pair and a negative pair. However, when both the normal image and the abnormal image are used to determine whether in-distribution or out-of-distribution in a 2D network, a class imbalance problem occurs. This is because the 2D network uses a slice of a volume as an input. A normal volume contains only normal slices, but an abnormal volume contains both normal slices and abnormal slices. In conclusion, normal slice images are overwhelmingly more than abnormal slice images, which hinders network training. Therefore, when computing the classification loss of the 2D network, binary cross entropy is calculated only for slices of abnormal volume. At this time, as shown in Figure \ref{imbalance}, there is an imbalance between the normal slice and the abnormal slice even within the abnormal volume. Therefore, in order to improve the learning performance, a weighted binary cross entropy loss that multiplies the abnormal slices by weight is used. This weight is multiplied only on the positive label, and the weight multiplies the ratio of the number of abnormal slices and the number of normal slices. This can be expressed as a formula as follows.
{\large $$L_{2d \, cls} = -(W_{i}*y_{i}\log{(p_{i})} + (1-y_{i})\log{(1-p_{i})}),$$
$$where\quad W_{i} = {Number\;of\;Normal\;Slices \over Number\;of\;Abnormal\;Slices}$$}

\vspace{1em}

\noindent \textbf{Label Smoothing. }Slice images used as input to a 2D network are labeled according to the presence or absence of anomalous pixels in each slice. In this case, even if the slice image contains very small anomalous pixels, such as the boundary between normal and abnormal in the figure \ref{imbalance}, it is also classified as an anomaly label. As the number of these input slices increases, it becomes difficult for the 2D network to properly learn classification. Therefore, when the pixels within 3\% of the entire slice pixel are abnormal, the label are changed into normal, and when the pixels within 3\% to 6\% of the slice are abnormal, the label are smoothed to 0.75. If more than 6\% of pixels are abnormal, the label is not changed.\\
Finally, adding the total loss of 2D network similarity loss between 3D networks and classification loss can be expressed as:
{\Large $$L_{2d \; total} = L_{2d \, cls} + \lambda_{1} L_{sim}$$}

\vspace{1.5em}
\subsection{Network Architecture}
\textbf{2D Network. }A deep 2D network such as ResNet-50 is carried out first using ImageNet-1k which is converted in grayscale. Since the increase in memory usage according to image size is not greater than that of the 3D volume, the 3D volume slice has twice as much resolution use as the input image in the 2D network. Through this, the advantage of using a higher resolution input image will also contribute to the improvement of accuracy.\\
\textbf{3D Network. }We used U-Net as a reference network that receives 3D patches as input~\cite{moodchallenge}. The segmentation task is performed by the decoder, and the classification task is performed by attaching a classification module to the bottom of the network.

\vspace{-.6em}
\section{Experiments} 

\noindent We divided MOOD data~\cite{mood_2021} into a training set, validation set, and test set as 8:1:1. During the training, apply the elastic deformation to input image with 50\% probability. 3D models were trained using Adam~\cite{adam} optimizer with one-cycle learning rate policy~\cite{onecycle} the range from $10^{-4}$ to $10^{-3}$ for 200 epochs in fine-tuning the pretask steps. For the 2D network, a ResNet-50 model as shown in Figure \ref{resnet50} was used. In the case of the brain, a 64 x 64 x 64 image was mostly usedn for 3D network and 128 x 128 sliced images which is twice the size of 3D network input image were used for 2D network. We set the different batch sizes 8, 2 for brainNet and AbdomNet considering the input image size. All of training was held on two NVIDIA RTX 3090 GPUs. \\
\indent We synthesize an OOD area from 10 pixel to half of the input size for each dimension. If an \textit{Copy-Paste} is generated in the background with the same intensity of air, we do not consider this area as a label. To make it possible, we preprocess the abdomen image to zero intensity outside the human body. After that, to prevent too small OOD, we perform the OOD synthesis process again when the generated OOD is less than 20 voxels. \\
\indent In the experiments, due to experiments on abdominal data take too much time using large-size images, only the verification will be made after finding the best method with brain data. \\

\begin{figure}[!t]
    \centerline{\includegraphics[width=17cm]{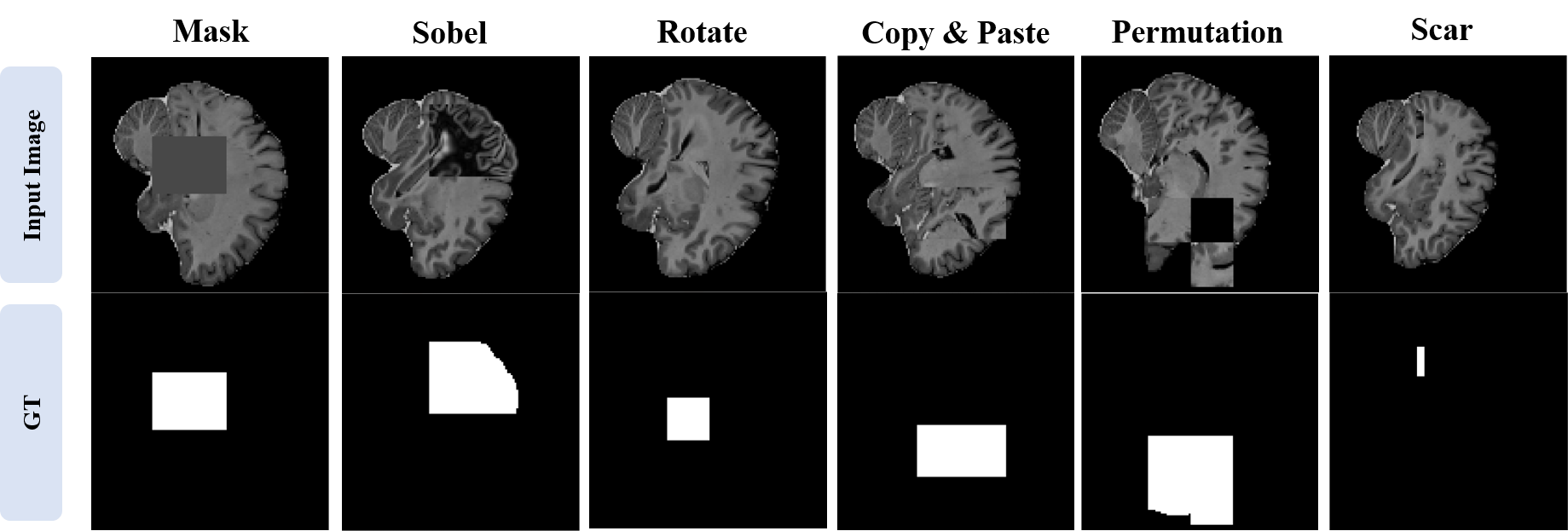}}
    \caption[Examples of Hard Augmentation.]{\textbf{Examples of Hard Augmentation}. The size and position of the patch to be augmented are randomly selected ~\cite{moodchallenge}.
    } \label{aug}
\end{figure}

\indent In order to create an reasonable test OOD dataset, we think of various types of augmentation that can be hard augmentation according to the fact that extreme augmentation ruins essential feature of the class~\cite{simclr,csi}. As in Figure \ref{aug}, there are six types of hard augmentation we have used: \textit{mask}, \textit{Sobel\_filter}, \textit{rotation}, \textit{Copy-Paste}, \textit{permutation}, and \textit{scar}. The \textit{mask} refers to a method of giving a random intensity value $\in$ $[0,1]$, the \textit{Sobel\_filter} refers to a method of applying a Sobel filter to a random patch, and \textit{rotation} refers to a method of rotating a random patch. The \textit{Copy-Paste} is a method of pasting a copied patch to another area, the \textit{permutation} is a method of mixing a random patch with 8 sub-patches divided in the x, y, and z-axis directions, and the \textit{scar}~\cite{scar} is a method of applying copy and paste to very small area. Among the parts to which augmentation is applied, if they have the same value as the existing image, the part is excluded from the OOD area. In case of the brain CT image, for example, if the background area is copied and pasted to another background area, this is excluded from the OOD area. During training, both normal and abnormal images for the same image are used as input images to learn features of the normal and the anomaly image simultaneously. \\
\indent Additionally, we tested on mask test dataset to check robustness to diverse shapes, positions, size and intensities. The mask test dataset have two shapes, twenty seven positions, three sizes, and four intensities of the OOD area. \\

\subsection{Pretask}
\noindent Tables \ref{pretask_result} shows the results of a hard augmentation test in which various pretasks are first trained on a 3D model and then anomaly detection is trained with fine-tuning way. When training pretasks, we use Adam~\cite{adam} optimizer with one-cycle learning rate policy~\cite{onecycle} the range from $10^{-3}$ to $10^{-2}$ for 100 epochs. Pretask training time takes less than one hours. For fine-tuning steps, anomaly detection training, we only use 3D network and train for 200 epochs. BrainNet~\cite{moodchallenge} refers to training with textit{Copy-Paste} for 100 epoch and then training with same method for 200 epochs again.\\

\begin{figure}[!t]
    \vspace{-2em}
    \centerline{\includegraphics[width=15cm]{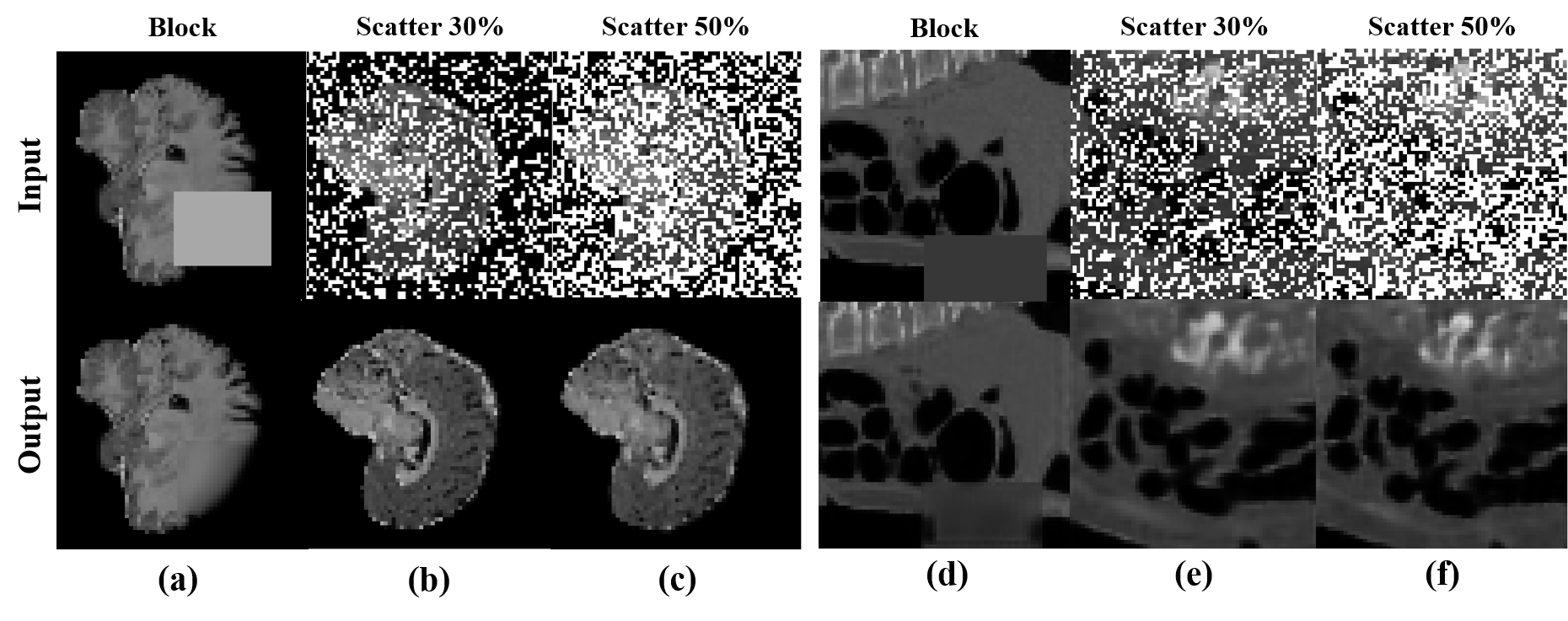}}
    \caption[Input and Prediction Results for Reconstruction.]{\textbf{Input and Prediction Results for Reconstruction.} Reconstruction is performed as pretasks. Then final results are got through fine-tuning.
    } \label{recon-result}
    \vspace{-1em}
\end{figure}

\indent Analyzing the results, it can be seen that both the segmentation and classification results of scar are greatly improved when pretrained with scatter-shaped reconstruction. Compared with the SoTA model, the AP greatly improved for anomaly with a small size and it shows that the accuracy also increased in most other cases. What is surprising is that there is no big difference between recovering 30\% of the area or 50\% of the area. Looking at the figure \ref{recon-result}, it can be seen that even in the case of a complex-shaped abdomen, the scattered shape is restored to the original shape, but the detail cannot be recovered at all in the case of the box shape. Through these results, it can be seen through the final performance improvement that recovering the original is helpful in learning the structure of the image. Also, the scattered reconstruction showed better results than the classification pretasks, which can be said to be helpful in the pretask using a decoder rather than the pretask using only the encoder in the encoder-decoder structure.\\

\begin{table*}[!t]
\caption[Segmentation and Classification Results on Six Types of Hard Augmentation Test for Pretasks.]{\textbf{Segmentation and Classification Results on Six Types of Hard Augmentation Test for Pretasks. }It shows fine-tuning results on various pretasks. Classification-based pretasks only utilize the encoder for training, while recontruction-based pretasks use the encoder and decoder simultaneously. }\label{pretask_result}
\vspace{1em}
\resizebox{\linewidth}{!}
{
\begin{tabular}[width=\textwidth]{ccccccccc}
\toprule
\multirow{2}{*}{\normalsize\bfseries Task} & \multirow{2}{*}{\normalsize\bfseries Pretask} & \multicolumn{6}{c}{\normalsize\bfseries Test} & \normalsize \multirow{2}{*}{\bfseries Total AP}\\
 & & Rotation & Scar & Sobel & Copy-Paste & Permutation & Mask & \\
\cmidrule(lr){1-1} \cmidrule(lr){2-2} \cmidrule(lr){3-8} \cmidrule(lr){9-9}
\multirow{7}{*}{\textbf{Seg}} & BrainNet~\cite{moodchallenge} & 0.9715 & 0.7737 & 0.9156 & \textbf{0.9942} & 0.9269 & 0.9922 & 0.9290 \\
\arrayrulecolor{Gray} \cmidrule(lr){2-9} \cmidrule(lr){2-9}
 & + Rotation & 0.9624 & 0.6175 & 0.9321 & 0.9904 & 0.9276 & 0.9848 & 0.9025 \\
 & + Jigsaw & 0.9660 & 0.7675 & 0.9163 & 0.9898 & 0.9275 & 0.9904 & 0.9262 \\
 & + Index & 0.8714 & 0.6741 & 0.8602 & 0.9265 & 0.8680 & 0.9034 & 0.8506 \\
 \arrayrulecolor{Gray} \cmidrule(lr){2-9} \cmidrule(lr){2-9}
 & + Box Recon & 0.9086 & 0.5089 & 0.8927 & 0.9587 & 0.8577 & 0.9470 & 0.8456 \\
 & + Scattered 30\% & \textbf{0.9750} & \textbf{0.8550} & \textbf{0.9432} & 0.9929 & \textbf{0.9399} & 0.9880 & \textbf{0.9490} \\
  & + Scattered 50\% & 0.9742 & 0.8542 & 0.9306 & 0.9936 & 0.9370 & \textbf{0.9940} & 0.9477 \\
  \arrayrulecolor{black} \hline
  \multirow{7}{*}{\textbf{Cls}} & BrainNet~\cite{moodchallenge} & 1.0 & 0.6125 & 1.0 & 1.0 & 0.9875 & 0.9875 & 0.9657 \\
  \arrayrulecolor{Gray} \cmidrule(lr){2-9} \cmidrule(lr){2-9}
 & + Rotation & 0.975 & 0.3625 & 0.9875 & 1.0 & 0.975 & 1.0 & 0.9354 \\
 & + Jigsaw & 0.9875 & 0.4375 & 0.9875 & 1.0 & 0.9875 & 0.9875 & 0.9427 \\
 & + Index & 1.0 & 0.7625 & 0.9375 & 1.0 & 0.9875 & 1.0 & 0.9677 \\
 \arrayrulecolor{Gray} \cmidrule(lr){2-9} \cmidrule(lr){2-9}
 & + Box Recon & 0.95 & 0.3875 & 0.9875 & 1.0 & 0.975 & 1.0 & 0.9167 \\
 & + Scattered 30\% & \textbf{1.0} & \textbf{0.7625} & 0.975 & 1.0 & 0.9875 & 0.9875 & \textbf{0.9675} \\
  & + Scattered 50\% & 0.9875 & 0.6625 & \textbf{0.9875} & 1.0 & 0.9875 & 0.9875 & \textbf{0.9677} \\
\arrayrulecolor{black} \toprule
\end{tabular}
}
\end{table*}

\begin{table*}[!t]
\caption[Results on Six Types of Hard Augmentation Test for Joint Embedding.]{\textbf{Results on Six Types of Hard Augmentation Test for Joint Embedding. }Scattered 30\% refers to BrainNet with reconstruction pretask without the joint embedding. Total AP is calculated by averaging the results of normal images and abnormal images.}\label{JE_result}
\vspace{1em}
\resizebox{\linewidth}{!}
{
\begin{tabular}[width=\textwidth]{ccccccccc}
\toprule
\multirow{2}{*}{\normalsize\bfseries Task} & \multirow{2}{*}{\normalsize\bfseries Method} & \multicolumn{6}{c}{\normalsize\bfseries Test} & \normalsize \multirow{2}{*}{\bfseries Total AP}\\
 & & Rotation & Scar & Sobel & Copy-Paste & Permutation & Mask & \\
\cmidrule(lr){1-1} \cmidrule(lr){2-2} \cmidrule(lr){3-8} \cmidrule(lr){9-9}
\multirow{4}{*}{\textbf{Seg}} & BrainNet & 0.9715 & 0.7737 & 0.9156 & 0.9942 & 0.9269 & 0.9922 & 0.9290 \\
& \textbf{Scattered 30\%} & 0.9750 & 0.8550 & 0.9432 & 0.9929 & 0.9399 & 0.9880 & 0.9490 \\
\arrayrulecolor{Gray} \cmidrule(lr){2-9} \cmidrule(lr){2-9}
 & JE (mean) & 0.9671 & 0.8088 & 0.9291 & 0.9925 & 0.9412 & 0.9436 & 0.9304 \\
 & \textbf{JE (max)} & 0.9767 & 0.8263 & 0.9545 & 0.9961 & 0.9605 & 0.9961 & \textbf{0.9517} \\
 \arrayrulecolor{black} \cmidrule(lr){1-9}
 \multirow{4}{*}{\textbf{Cls}} & BrainNet & 1.0 & 0.6125 & 1.0 & 1.0 & 0.9875 & 0.9875 & 0.9657 \\
& \textbf{Scattered 30\%} & 1.0 & 0.7125 & 0.9875 & 1.0 & 0.9875 & 0.9875 & 0.9610 \\
\arrayrulecolor{Gray} \cmidrule(lr){2-9} \cmidrule(lr){2-9}
 & JE (mean) & 0.9625 & 0.5200 & 0.9875 & 1.0 & 0.9750 & 1.0 & 0.9138 \\
 & \textbf{JE (max)} & 0.9625 & 0.8500 & 1.0 & 1.0 & 0.9875 & 1.0 & \textbf{0.9771} \\
\arrayrulecolor{black} \toprule
\end{tabular}
}
\end{table*}

\vspace{1em}
\subsection{Joint Embedding}
\begin{figure}[!h]
    \leftline{\includegraphics[width=8cm]{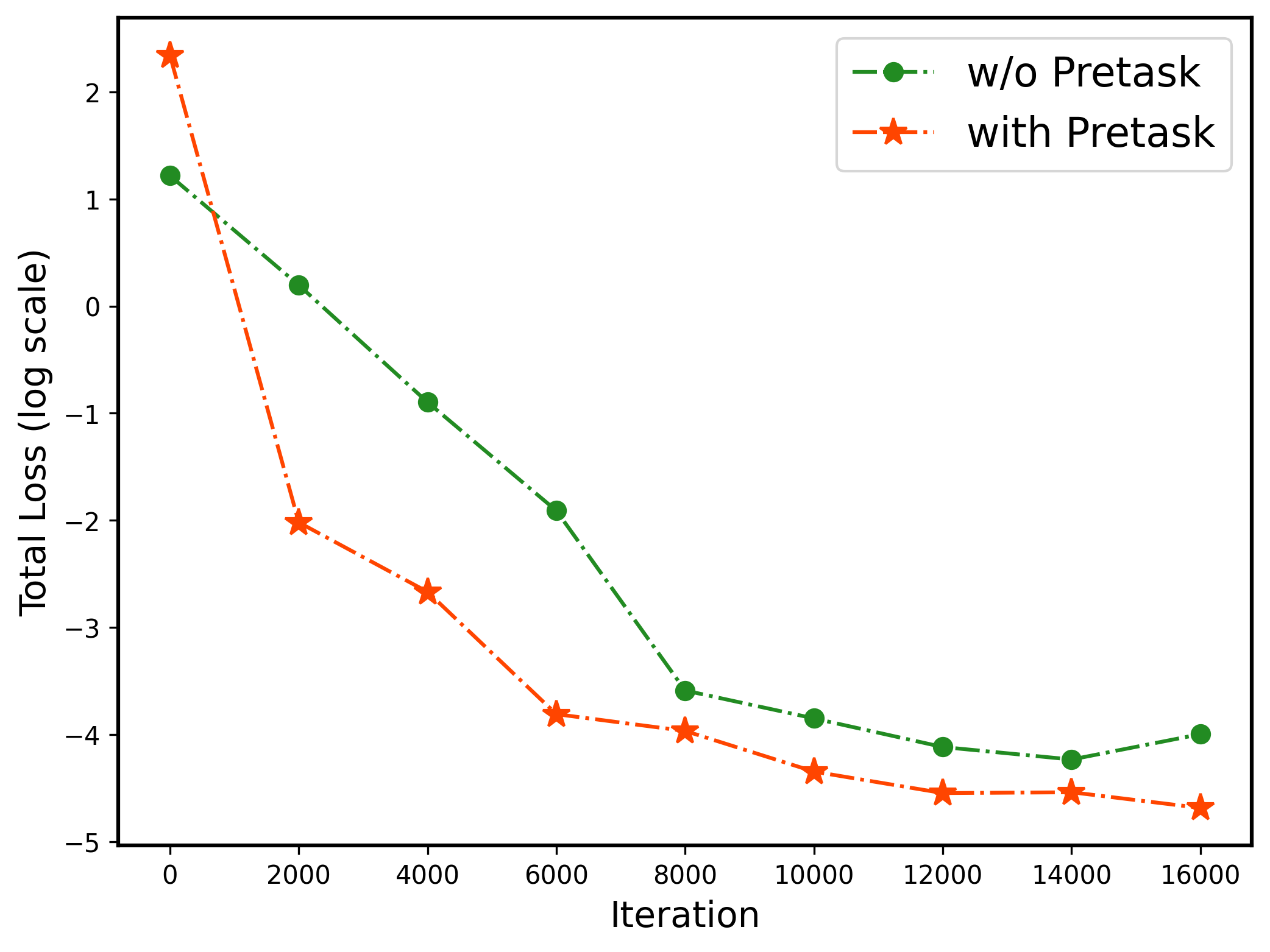}
    \includegraphics[width=8cm]{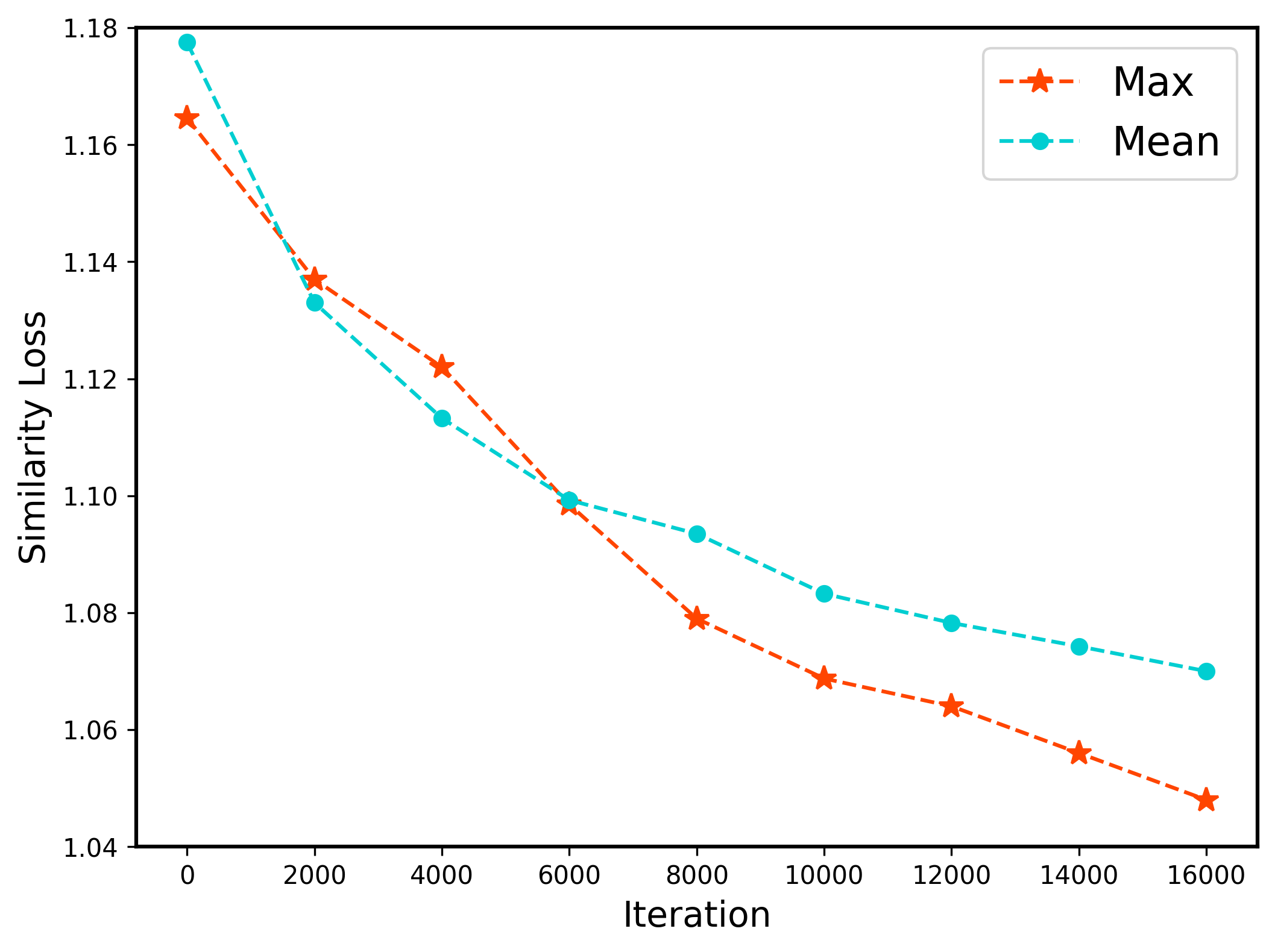}}
    \caption[Comparison of Similarity Loss between Mean Shape Reduction and Max Shape Reduction.]{\textbf{Comparison of (Left) SoTA method and With Scattered Reconstruction Pretask (Right) Similarity Loss between Mean Shape Reduction and Max Shape Reduction. }w/o Pretask means the BrainNet that does not utilize any pretask and with pretask means training while fine-tuning the scattered reconstruction pretask. Mean shape reduction refers to reduce the 2D latent vector shape by averaging each section and max shape reduction refers to reduce the 2D latent vector shape by picking maximum values of each section.
    } \label{plot}
\end{figure}

\noindent An experiment was conducted on ResNet-50 with gray-scale ImageNet-1k pretrained model for 2D network. For the 3D network, the BrainNet with the pretask of scattered 30\% fine-tuning, which showed the best performance in Section 4.1 is used. Both networks compare their latent vectors for every iteration through similarity loss.\\
\indent According to Table \ref{JE_result}, it can be seen that the classification AP of scar significantly increases. Method written in scattered 30\% means that fine-tuning results of reconstruction method without joint embedding. It can be seen in Figure ~\ref{plot} that allowing the 3D network to learn the representation by a pretask increases the learning speed compared to training from scratch. When the pretask is not performed, the loss decreased slowly, whereas when the pretask is performed, the loss decreased relatively quickly. This tendency confirms that it contributed to increasing the initial training speed. \\
\indent The results of applying 2D network and joint embedding during training were better in both segmentation and classification than the results of fine-tuning only after the pretask on BrainNet. This means that the 2D deep network and pretrained with large-capacity images are helpful for the 3D model. Also, in the case of a 2D network, it is thought that it can detect a small anomaly area more accurately because it receives an image with twice the resolution as an input. However, the 3D model, the network for classification is attached in the form of a module. Therefore, it can be analyzed that the classification AP increase as the 2D network compensates for these shortcomings. The segmentation also increase the AP overall, which refers that the similarity loss encourage training of the 3D network overall. \\
\indent In order to obtain the similarity loss, the representation of the 2D network is reduced to have the same shape as the representation of the 3D network. In the process, experiments are also conducted on whether to extract features with average values or features with maximum values based on the axis to be reduced. As the results in Table \ref{JE_result}, it can be seen that extracting features with maximum value shows a positive effect than when extracting features with mean value. This indicates that joint embedding works appropriately when the most important features for each region extracted from the 2D network are compared with the features extracted from the 3D network. This is because, in the case of average-valued features, important features may be diluted. As shown in Figure ~\ref{plot}, when shape reduction is performed with the average value, the similarity loss quickly converges to a certain value, but in the case of shape reduction with the maximum value, the loss continues to decrease, suggesting that learning continues. \\

\section{Ablation Study}

\noindent An ablation study is performed on the models under the same condition that they are trained for 200 epochs. First, an ablation study on pretrained weights is conducted. We check whether the accuracy of the 3D model can be improved by the huge data called gray-scale ImageNet-1k in the 2D network. Looking at the table \ref{ablation}, when the 2D network is not pretrained, the segmentation and classification performance is slightly degraded. It can be seen that the 2D network with an encoder that extracts important features through the process of pretraining with a large gray-scale dataset helped to improve the performance of the 3D network. Comparing the 3D network with and without pretask, the segmentation AP dropped significantly. Joint embedding has the disadvantage that it takes a lot of time to go from scratch, so it does not seem to converge yet. Considering that it takes about one hour to train the pretask, it can be confirmed that performing the pretask first and then anomaly detection as the second stage causes fast joint embedding. \\
In addition, we analyze the results of an ablation study on the weighted binary cross entropy and label smoothing introduced to solve the class imbalance problem in the 2D network. If the weighted binary cross entropy is not used, it is often judged to be normal even though it is an anomaly image because it focuses a lot on normal. Through this, it can be said that the weighted binary cross entropy contributed to solving the class imbalance problem to some extent. Finally, in the case of label smoothing, it causes some performance improvement. These results greatly improved the performance of the 2D network, which means that it have a significant impact on the improvement of the classification performance of the 3D network. \\

\begin{table*}[!t]
\caption[Results for Ablation Study.]{\textbf{Results for Ablation Study.} 2D network is pretrained with gray-scale ImageNet-1k dataset, and 3D network conduct the scattered shape reconstruction pretask. WBCE means weight binary cross entropy loss which solves class imbalance problem on 2D network. Label smoothing is also applied to 2D network to alleviate too difficult classification tasks.}\label{ablation}
\vspace{1em}
\resizebox{\linewidth}{!}
{
\begin{tabular}[width=\textwidth]{cccccc}
\toprule
 \multicolumn{4}{c}{\normalsize\bfseries Method} & \multicolumn{2}{c}{\bfseries Total AP}\\
 2D Network Pretraining & 3D Network Pretask & WBCE & Label Smoothing & Segmentation & Classification \\
\cmidrule(lr){1-4} \cmidrule(lr){5-6}
\color{red}\xmark & \color{ForestGreen}\cmark & \color{ForestGreen}\cmark & \color{ForestGreen}\cmark & 0.9495 & 0.9594 \\
\color{ForestGreen}\cmark & \color{red}\xmark & \color{ForestGreen}\cmark & \color{ForestGreen}\cmark & 0.7232 & \textbf{0.9721} \\
\color{ForestGreen}\cmark & \color{ForestGreen}\cmark & \color{red}\xmark & \color{ForestGreen}\cmark & \textbf{0.9534} & 0.9276 \\
\color{ForestGreen}\cmark & \color{ForestGreen}\cmark & \color{ForestGreen}\cmark & \color{red}\xmark & 0.9475 & 0.9428 \\
\color{ForestGreen}\cmark & \color{ForestGreen}\cmark & \color{ForestGreen}\cmark & \color{ForestGreen}\cmark & \textbf{0.9518} & \textbf{0.9702}\\
\toprule
\end{tabular}
}
\end{table*}

\vspace{1em}
\section{CONCLUSION} \label{sec:latex}
\noindent Although the performance of the 3D network shows overwhelming capability in that it learns three dimensional contexts, there are many limitations in making a deep network due to the GPU memory problem. Since anomaly detection has to perform classification and segmentation at the same time, the memory available for each task is inevitably more limited. On the other hand, 2D networks have less memory restrictions to create deep networks, so various deep and wide networks exist. In addition, from the perspective of the lack of datasets, a chronic problem in the medical image field, the 2D network has the advantage of being able to utilize various and large-capacity natural image datasets. As such, the 2D network and the 3D network have strength points of complementing each other, so they can benefit from online learning together. By comparing the similarity of features extracted from each network, our method induces similar features to be aligned in the embedding space. In this process, we show that the 2D network and the 3D network together train to improve each other's performance. In this thesis, we also propose a method of pretraining a 2D network through a large amount of natural images and training a pretask called scattered-shape reconstruction in a 3D network first so that joint embedding can proceed quickly. These steps warm up each network so that it can extract important representations before the main training. In this way, we prove that the main training proceeds more time-efficiently by conducting the training in two stages. The main contribution of the thesis is that we present a new approach of applying joint embedding to networks of different dimensions through similarity comparison. Our method gives superior performance than the existing SoTA model on the MOOD 2021 dataset, a medical anomaly detection dataset. Finally, on other datasets, our method also shows higher performance than learning the 3D network alone, demonstrating the sufficient possibility that it can be used in practice.\\


\bibliographystyle{unsrt}  
\bibliography{main}

\end{document}